\documentclass[10pt,twocolumn,letterpaper]{article}

\usepackage[pagenumbers]{iccv} 

\usepackage{multirow}
\usepackage{dsfont}
\usepackage{wasysym}
\usepackage{arydshln}

\definecolor{iccvblue}{rgb}{0.21,0.49,0.74}
\usepackage[pagebackref,breaklinks,colorlinks,allcolors=iccvblue]{hyperref}

\def\paperID{***} 
\def\confName{ICCV}
\def\confYear{2025}

\title{MTA: Multimodal Task Alignment for BEV Perception and Captioning}

\author{
Yunsheng Ma$^{1,2}$\thanks{Work done while interned at Bosch Research North America. \newline \ \ \ \ \ \indent \ $\dagger$  Corresponding author.}  \ \  
Burhaneddin Yaman$^{1\dagger}$ \ 
Xin Ye$^{1}$ \\
Jingru Luo$^{1}$ \
Feng Tao$^{1}$\
Abhirup Mallik$^{1}$ \ 
Ziran Wang$^{2}$ \ 
Liu Ren$^{1}$\\
$^{1}$Bosch Research North America \& Bosch Center for Artificial Intelligence (BCAI) \ \ \ \\ 
$^{2}$Purdue University\\
{\tt\small \{yunsheng,ziran\}@purdue.edu} \ \ \ \\  {\tt\small \{burhaneddin.yaman,xin.ye3,jingru.luo,feng.tao2,abhirup.mallik,liu.ren\}@us.bosch.com}
}

\begin{document}
\maketitle
\begin{abstract}
Bird's eye view (BEV)-based 3D perception plays a crucial role in autonomous driving applications. The rise of large language models has spurred interest in BEV-based captioning to understand object behavior in the surrounding environment. However, existing approaches treat perception and captioning as separate tasks, focusing on the performance of only one task and overlooking the potential benefits of multimodal alignment. To bridge this gap between modalities, we introduce MTA, a novel multimodal task alignment framework that boosts both BEV perception and captioning. MTA consists of two key components: (1) BEV-Language Alignment (BLA), a contextual learning mechanism that aligns the BEV scene representations with ground-truth language representations, and (2) Detection-Captioning Alignment (DCA), a cross-modal prompting mechanism that aligns detection and captioning outputs. MTA seamlessly integrates into state-of-the-art baselines during training, adding no extra computational complexity at runtime. Extensive experiments on the nuScenes and TOD3Cap datasets show that MTA significantly outperforms state-of-the-art baselines in both tasks, achieving a $10.7\%$ improvement in challenging rare perception scenarios and a $9.2\%$ improvement in captioning. These results underscore the effectiveness of unified alignment in reconciling BEV-based perception and captioning.
\end{abstract} 
\section{Introduction}
\label{sec:intro}

3D perception is a fundamental and crucial task for embodied AI applications
such as robotics and autonomous driving
\cite{chen_end--end_2024,jia_bench2drive_2024}.
Among 3D perception methods, bird's eye view (BEV)-based methods have recently
gained significant attention, particularly in the context of autonomous driving tasks
\cite{hu_st-p3_2022,hu_planning-oriented_2023,jiang_vad_2023,weng_para-drive_2024}.
Unlike monocular frameworks which process each camera view separately, BEV provides
a unified representation of a given scene by fusing information from multi-view camera
images or other sensory inputs such as LiDAR scans~\cite{li_bevformer_2022,liu_bevfusion_2023}.
The generated BEV representations serve as the primary source of information for
solving downstream autonomous driving tasks such as detection and tracking.

In recent years, transformer-based BEV methods have shown rapid progress,
enabling the extraction of spatio-temporally holistic representations of the
surrounding environment from multi-view camera images~\cite{li_bevformer_2022,yang_bevformer_2023}.
These rich representations have facilitated achieving state-of-the-art 3D perception
performance. The rise of foundation models, such as multimodal large language models
(MLLMs) has led to the emergence of research on explainability and understanding
of 3D scenes~\cite{ma_position_2025,xu_drivegpt4_2024,marcu_lingoqa_2024,sima_drivelm_2024}. This
task is manifested as a captioning task which aims to describe the localization,
context, and behavior of objects in the scene in the form of natural language. 3D
captioning has been extensively investigated for various indoor applications~\cite{chen_scan2cap_2021,chen_vote2cap-detr_2024}
and has more recently been extended to outdoor applications such as perception task
in autonomous driving~\cite{jin_tod3cap_2024}. BEV-based 3D captioning extracts
information from BEV and task heads such as 3D detection and uses it as the
condition for caption generation.

While there is an increasing number of research studies on BEV perception and captioning
tasks, the joint alignment between modalities, which aims to enhance the performance
of both modality tasks, has not been properly addressed. In particular, one
stream of works focuses on BEV-based detection without considering the
captioning performance~\cite{li_bevformer_2022}, while another stream of works focuses on captioning performance
without reporting performance on perception tasks such as 3D detection~\cite{jin_tod3cap_2024}. However,
these two tasks are not disjoint and can complement each other by enforcing multimodal
alignment strategies that have the potential to significantly advance the field of
3D perception and captioning in autonomous driving applications.

To bridge this gap, we introduce MTA, a multimodal task alignment approach for
BEV perception and captioning. The proposed MTA approach presents two mechanisms
for alignment, namely BEV-Language Alignment (BLA) and Detection-Captioning
Alignment (DCA). BLA introduces a multimodal contextual learning mechanism that incorporates ground-truth
caption representations to learn the alignment between the BEV visual representations
of the scene and the natural language-based scene understanding. Rather than
relying solely on the language modeling objective, BLA provides additional supervision
by aligning the BEV-based contextual object queries with corresponding ground-truth
linguistic representations obtained from pretrained text encoders. DCA, on the other
hand, aims to explicitly promote consistency between the perception outputs from
the visual branch and captioning outputs from the language branch. It introduces a cross-modal
prompting mechanism that encourages the MLLM to generate captions that are coherent
with the predicted bounding boxes and class labels. DCA goes beyond relying
solely on the gradients from single modality objectives (detection loss or language
modeling loss) to optimize the task heads or the MLLM.

MTA is a flexible framework that can be seamlessly integrated into existing BEV-based
perception and captioning frameworks. Additionally, the proposed MTA modules are
only used during training time to enforce alignment between modalities. Thus,
MTA does not require any architectural changes or introduce any additional
computational overhead during inference time, which is very critical for downstream
tasks such as autonomous driving. 
In this paper, we evaluate the importance of MTA on the state-of-the-art
frameworks using the challenging large-scale nuScenes~\cite{caesar_nuscenes_2020} and TOD3Cap datasets~\cite{jin_tod3cap_2024}. Experimental
results show that MTA outperforms the previous best baseline on both perception and
captioning tasks. In particular, MTA provides an improvement of 4.9\% and 9.2\% in
terms of perception and captioning metrics, respectively, over the corresponding
state-of-the-art baseline. Moreover, qualitative results further confirm the quantitative
findings, demonstrating that MTA not only achieves superior performance metrics
but also reduces the occurrence of hallucinated captions, which is a fundamental
factor for safety-critical applications like autonomous driving. Our main contributions can be summarized as follows:
\begin{itemize}
    \item We propose MTA, a novel multimodal task alignment framework that
        bridges the gap between BEV-based perception and captioning tasks.

    \item MTA introduces two novel alignment modules, BEV-Language Alignment (BLA)
        and Detection-Captioning Alignment (DCA), which enforce alignment through multimodal
        contextual learning and cross-modal prompting mechanisms, respectively.

    \item MTA seamlessly integrates into existing architectures and does not introduce
        any additional computational overhead during inference, as both of the
        MTA components are only active during training.

    \item Extensive experiments on the challenging nuScenes and TOD3Cap datasets
        demonstrate that MTA consistently outperforms state-of-the-art methods and
        in both perception and captioning tasks.
\end{itemize}
\section{Related Work}
\label{sec:related_work}

\subsection{BEV Perception}
In recent years, BEV frameworks have utilized transformer architectures that
generate high-quality BEV feature maps~\cite{li_bevformer_2022,yang_bevformer_2023,liang_bevfusion_2022,liu_bevfusion_2023,xu_cobevt_2022,ye_bevdiffuser_2025}.
Among these works, BEVFormer has unraveled a new era in BEV perception by fusing
information from multi-view camera images both spatially and temporally to
acquire a spatio-temporally holistic representation of the scene~\cite{li_bevformer_2022}.
Another notable work in this domain is BEVFusion, which presents a framework to
fuse BEV feature maps from both camera and LiDAR sensors for efficient and
robust BEV perception~\cite{liu_bevfusion_2023}. These advancements in BEV perception
have laid the foundation for a more comprehensive understanding of 3D environments
in autonomous driving applications.

\subsection{3D Captioning}
3D captioning aims to provide natural language descriptions of the localization and
behavior of objects in a given scene. Recently, the field of 3D captioning has witnessed
significant progress, thanks to the rapid emergence of multimodal large language
models and the release of many public datasets, primarily for indoor applications~\cite{chen_scan2cap_2021,chen_end--end_2023,chen_vote2cap-detr_2024}.
These advancements have prompted the embodied AI community to collect 3D
captioning datasets and develop 3D captioning frameworks for outdoor
applications such as autonomous driving~\cite{malla_drama_2023,wang_omnidrive_2024,qian_nuscenes-qa_2024}. A notable work in this direction is TOD3Cap~\cite{jin_tod3cap_2024},
which has released a large captioning dataset for autonomous driving and
proposed a framework for BEV-based 3D dense captioning. This framework utilizes information
from BEV and 3D perception outputs as inputs to an MLLM for generating captions.

Despite these advancements in BEV perception and 3D captioning, there remains a
significant gap in jointly optimizing and aligning these two modalities to
enhance the performance of both tasks, which we aim to address in this study through
the proposed MTA framework. 

\begin{figure*}
    \centering
    \includegraphics[width=1.0\linewidth]{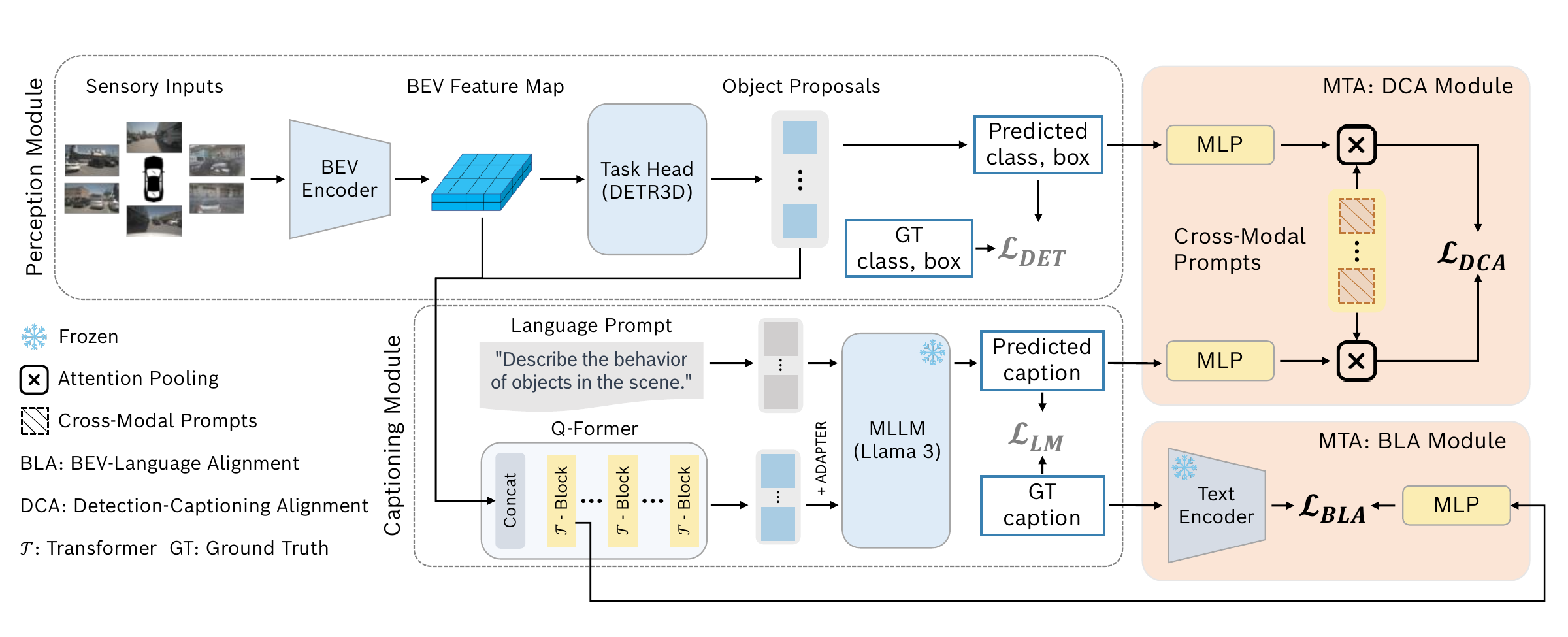}
    \caption{\textbf{Overview of proposed multimodal task alignment (MTA) framework.} MTA
    enables joint performance improvement for BEV perception and dense captioning
    tasks through BEV-Language Alignment (BLA) and Detection-Captioning Alignment (DCA)
    mechanisms. In particular, BLA is a contextual learning mechanism to reconcile
    BEV scene representation with language-based scene understanding, and DCA is
    a cross-modal prompting mechanism to promote consistency between detection
    and captioning outputs. The MTA is trained end-to-end with a combination of task-specific
    losses (detection and captioning losses), along with the BLA and DCA
    objectives. }
    \label{fig:mta}
\end{figure*}

\subsection{Vision-Language Models}
Vision-language models, trained on massive internet-scale data, have shown
strong promise in learning good representations for downstream tasks. Pioneering
works such as CLIP~\cite{radford_learning_2021}, ALIGN~\cite{jia_scaling_2021}, and
Florence~\cite{yuan_florence_2021,xiao_florence-2_2024} pretrain multimodal models
on millions to billions of image-text pairs, showing strong zero-shot performance on various
tasks like image classification and retrieval. Subsequent works have proposed strategies,
such as prompting to efficiently adapt VLMs to new domains and datasets
\cite{zhou_learning_2022,khattak_maple_2023}. More recently, VLMs have been
explored in BEV-based autonomous driving applications, where they have been either
used for scene understanding \cite{mu_most_2024,cao_maplm_2024,liang_aide_2024} or for improving the performance
of autonomous driving tasks such as perception and planning \cite{tian_tokenize_2024,tian_drivevlm_2024,mao_language_2024,hwang_emma_2024,li_driving_2024}.
Unlike previous methods, MTA emphasizes improving alignment between vision and language
modalities to jointly enhance BEV perception and captioning tasks.

\section{Methodology}
\label{sec:methodology} \textbf{Overview.} The overall framework of multimodal task
alignment (MTA), which strengthens the alignment between BEV perception and
captioning tasks to achieve state-of-the-art performance on both tasks, is illustrated
in \cref{fig:mta}. The methodology section is outlined as follows. In \cref{sec:prelim},
we provide a background on BEV perception and captioning tasks. In \cref{sec:bla}
and \cref{sec:dca}, we provide details of the proposed MTA alignment mechanisms,
namely BEV-Language Alignment and Detection-Captioning Alignment. Finally,
in \cref{sec:training}, we provide the overall loss function for training the MTA
framework.

\subsection{Preliminaries}
\label{sec:prelim}

\paragraph{BEV Perception Module.}
The BEV perception module $\mathcal{D}$ processes sensory inputs such as camera,
LiDAR, or both, to obtain a unified top-down representation of the surrounding environment.
In the context of given camera sensors, multi-view camera images are processed through
a backbone to obtain multi-view camera features. Subsequently, the resulting
perspective-view features are fed into BEV encoders such as BEVformer, which
lift these image features into BEV space by spatio-temporally fusing them~\cite{li_bevformer_2022}.

Subsequently, the generated BEV feature maps are fed into a downstream task head,
such as a transformer decoder for 3D detection~\cite{wang_detr3d_2021}. Due to a
lack of ground-truth BEV maps, BEV perception is trained end-to-end with the
objective of minimizing the task head loss function. The evaluation performance of
the task head serves as a proxy for the quality of BEV perception.

\paragraph{BEV Captioning Module.}
The BEV captioning module $\mathcal{G}$ aims to generate natural language
descriptions of the localization and behavior of objects in the scene. It takes the
BEV perception outputs, such as the BEV feature map and object proposals from
the task head, as input. A relation query transformer (Q-Former) is generally
employed to extract and transfer contextual information from the BEV perception
to language space \cite{li_blip-2_2023,wang_omnidrive_2024,jin_tod3cap_2024}. Formally, the Q-Former maps the embedding for each detected object to the language space
as follows:
\begin{equation}
    \mathbf{d}_{i} = \mathcal{Q}_{i}\left([\mathbf{d}_{i-1}, F_{i-1}^{\text{BEV}}
    ]\right) \quad i=1,2,\cdots,L
\end{equation}
where $[\cdot,\cdot]$ denotes the concatenation operation, $\mathbf{d}_{0}$
represents the detection embeddings from the detection head, $F_{0}^{\text{BEV}}$
is the BEV feature map from the BEV encoder, $L$ is the total number of
transformer blocks in the Q-Former. The refined object queries are then projected
into the latent dimension of the MLLM using a multilayer perceptron (MLP):
$\mathbf{q}=\Phi^{\text{L}}(\mathbf{d}_{L})$, where $\mathbf{q}$ denotes the projected queries and $\Phi^{\text{L}}$ is the MLP. The MLLM takes the projected queries
$\mathbf{q}$ and a language prompt $\mathbf{p}$ as input and generates captions for
each object. Formally, the MLLM models the conditional probability of generating
a language output sequence $\mathbf{o}$ given the multimodal input:
\begin{equation}
    \Pr(\mathbf{o}|\mathbf{p},\mathbf{q})=\prod_{i=1}^{n}\Pr(o_{i}|o_{<i},\mathbf{p}
    ,\mathbf{q})
\end{equation}
where $n$ is the number of output tokens, and $o_{<i}$ denotes the previous tokens.
The captioning module is efficiently trained by minimizing the language modeling
loss $\mathcal{L}_{LM}$ using adapters~\cite{zhang_llama-adapter_2024}.

\subsection{BEV-Language Alignment}
\label{sec:bla} Our goal is to bridge the gap between the BEV-based scene representation
used for 3D detection and the language-based scene understanding and reasoning capabilities
of the MLLM. However, off-the-shelf MLLMs cannot directly comprehend and reason on
BEV features, as they have not been exposed to such representations during their
pretraining phase. Furthermore, the alignment gap between BEV features and the
MLLM's language space is more pronounced compared to the visual tokens used in
general-domain MLLMs~\cite{chen_internvl_2024}.

To address this challenge, we introduce a novel BEV-Language Alignment (BLA)
module that explicitly aligns the BEV perception features, which encompass visual
contextual information for each object, with their corresponding ground-truth
linguistic representations. By aligning the Q-Former's visual BEV features with ground-truth captioning features, we strengthen the alignment between the BEV
perception and captioning modules, enabling the MLLM to better comprehend and reason
over the BEV representation.

Formally, the BLA module operates as follows. Given a ground-truth caption $\mathbf{o}$,
we compute its text embedding using a pretrained CLIP text encoder $\mathcal{T}$~\cite{radford_learning_2021}.
We then extract the projected Q-Former features $\Phi^{\text{Q}}(\mathbf{q}_{\ell}
)$, where $\mathbf{q}_{\ell}$ represents the hidden states from the $\ell$-th
layer of the Q-Former, and $\Phi^{\text{Q}}$ is a trainable projection head parameterized
as an MLP. The alignment is enforced using a mean squared error loss formulated
as:

\begin{equation}
    \mathcal{L}_{\text{BLA}}(\Phi^{\text{Q}},\mathcal{Q}_{1:\ell})=\left\|\Phi^{\text{Q}}
    (\mathbf{q}_{\ell})-\mathcal{T}(\mathbf{o})\right\|_{2}^{2}.
\end{equation}

Conceptually, the BLA-enhanced Q-Former can be viewed as a two-stage process. In
the first stage (before the $\ell$-th layer), the Q-Former focuses on learning a
context-aware representation of the object queries by attending to the BEV
features. This stage allows the Q-Former to capture the spatio-temporal relationships
and semantics encoded in the BEV representation with direct representation-based
supervision. In the second stage (from the $\ell$-th layer onwards), the Q-Former
maps the object query features into an MLLM-aligned space, making them more
amenable to the MLLM's language-based reasoning capabilities.

\subsection{Detection-Captioning Alignment}
\label{sec:dca} In current BEV-based perception and captioning frameworks, the 3D
detection and captioning tasks are typically optimized independently, which may
lead to suboptimal performance and a lack of coherence between the predicted
bounding boxes and the generated captions. To address this limitation, we
further propose a Detection-Captioning Alignment (DCA) module that aims to bridge
the gap between the detection and captioning outputs. The main challenge here lies
in the significant discrepancy between the modalities of the detection labels (class
labels and bounding box coordinates) and the captioning logits (language tokens).
Directly aligning these outputs can lead to a performance drop in both tasks.

We tackle this challenge by introducing a cross-modal prompting approach. We define
a set of $N$ learnable prompt tokens $\mathcal{P}= \left\{\mathbf{p}_{1},...,\mathbf{p}
_{N}\right\}\in\mathbb{R}^{N\times D}$ that serves as a shared embedding space for
aligning the detection and captioning outputs. Formally, let $\hat{\mathbf{c}}$ and
$\hat{\mathbf{b}}$ denote the class labels and bounding box coordinates from the
detection head, respectively. We project the concatenated detection outputs into
the cross-modal prompt space via attention pooling:
\begin{align}
    \mathbf{x}_{\text{det}} & =\left[\Phi^{\text{cls}}(\hat{\mathbf{c}}),\Phi^{\text{box}}(\hat{\mathbf{b}})\right]                                                                                                           \\
    \mathbf{p}_{\text{det}} & =\sum_{i=1}^{N}\left(\frac{\exp\left(\mathbf{x}_{\text{det}}^{\top}\cdot\mathbf{p}_{i}\right)}{\sum_{j=1}^{N}\exp\left(\mathbf{x}_{\text{det}}^{\top}\cdot\mathbf{p}_{j}\right)}\right)\mathbf{p}_{i}
\end{align}
where $\Phi^{(\cdot)}$ denotes trainable projection heads parameterized as MLPs.

Similarly, we project the captioning logits into the same prompt space:
\begin{equation}
    \mathbf{p}_{\text{cap}}=\sum_{i=1}^{N}\left(\frac{\exp\left(\left(\Phi^{\text{cap}}(\hat{\mathbf{o}})\right)^{\top}\cdot\mathbf{p}_{i}\right)}{\sum_{j=1}^{N}\exp\left(\left(\Phi^{\text{cap}}(\hat{\mathbf{o}})\right)^{\top}\cdot\mathbf{p}_{j}\right)}
    \right)\mathbf{p}_{i}
\end{equation}
where $\hat{\mathbf{o}}$ represents the captioning logits from the MLLM.

\noindent
Finally, to enforce alignment between the prompt-aligned detection and captioning
embeddings, we employ the CLIP contrastive loss~\cite{radford_learning_2021}:

\begin{equation}
    \mathcal{L}_{\text{DCA}}(\mathcal{D}, \mathcal{G}) = \mathcal{L}_{\text{CLIP}}
    \left(\mathbf{p}_{\text{det}},\mathbf{p}_{\text{cap}}\right).
\end{equation}

\noindent
By minimizing $\mathcal{L}_{\text{DCA}}$, we encourage the detection and
captioning outputs to be aligned in the shared prompt space.

The DCA module enhances the BLA module by explicitly enforcing consistency
between the primary outputs of the perception and captioning branches. By incorporating
both alignment mechanisms, our proposed framework attains a more comprehensive
understanding of the 3D scene and enables more accurate caption generation
grounded in the detected objects.

\subsection{Training}
\label{sec:training} The final loss function for training the proposed MTA
framework is a weighted combination of the detection loss
$\mathcal{L}_{\text{DET}}$, the language modeling loss $\mathcal{L}_{\text{LM}}$,
the BEV-language alignment loss $\mathcal{L}_{\text{BLA}}$, and the detection-captioning
alignment loss $\mathcal{L}_{\text{DCA}}$:
\begin{equation}
    \mathcal{L}_{\text{MTA}}= \alpha\mathcal{L}_{\text{DET}}+ \beta\mathcal{L}_{\text{LM}}
    + \lambda_{1}\mathcal{L}_{\text{BLA}}+ \lambda_{2}\mathcal{L}_{\text{DCA}}
\end{equation}

By default, we do not tune and set $(\alpha,\beta)=(10,1)$ following~\cite{jin_tod3cap_2024},
and set $(\lambda_{1},\lambda_{2})=(1,10^{-2})$ to ensure a balanced magnitude.
\section{Experiments and Results}
\label{sec:experiments} We conduct a comprehensive evaluation of the proposed
MTA framework, demonstrating its effectiveness in improving both 3D dense captioning
and detection performance through novel alignment mechanisms. The experimental
setup, including datasets, evaluation metrics, and implementation details, is
described in \cref{sec:setup}. In \cref{sec:results}, we compare MTA's
performance against the baseline TOD3Cap network and other state-of-the-art
methods, along with qualitative results. Finally, \cref{sec:ablation} presents ablation studies to validate MTA's alignment components.

\subsection{Experimental Set-up}
\label{sec:setup}
\paragraph{Datasets.}
We conduct comprehensive experiments on the nuScenes~\cite{caesar_nuscenes_2020}
and TOD3Cap~\cite{jin_tod3cap_2024} datasets. NuScenes is a widely used
benchmark in autonomous driving, containing 700 training and 150 validation
scenes. Each scene is captured for a duration of $\sim$20s using six cameras
covering the entire 360-degree field of view, with key samples annotated at 2Hz. The detection task contains 1.4M annotated bounding boxes from 10 object
categories. The TOD3Cap dataset extends nuScenes with dense language captioning
annotations, providing approximately 2.3M language descriptions, with an
average of 2.7K descriptions per scene.
\vspace{-5pt}
\paragraph{Perception Metrics.}
For the BEV perception task, we report the standard 3D object detection metrics
within the nuScenes dataset, including: mean Average Precision (mAP), Average Translation
Error (ATE), Average Scale Error (ASE), Average Orientation Error (AOE), Average
Velocity Error (AVE), Average Attribute Error (AAE), and nuScenes Detection
Score (NDS). More details on detection metrics can be found in~\cite{caesar_nuscenes_2020}.
The reported results are calculated using the validation split for all
experiments.

\paragraph{Captioning Metrics.}
For the BEV captioning task, we report the $m\text{@IoU=}k$
metric~\cite{chen_scan2cap_2021}. Let $(\mathbf{b}_{i}, \mathbf{o}_{i})$ denote each ground-truth
box-caption pair, where $\mathbf{b}_{i}$ and $\mathbf{o}_{i}$ are
the bounding box coordinates and the caption for the $i$-th object, respectively.
The predicted box-caption pair is denoted as
$(\hat{\mathbf{b}}_{i}, \hat{\mathbf{o}}_{i})$. The $m\text{@IoU=}k$ metric is
formulated as:

\begin{equation*}
    m\text{@IoU=}k := \frac{1}{N}\sum_{i=1}^{N}m(\hat{\mathbf{o}}_{i}, \mathbf{o}_{i}
    ) \cdot \mathds{1}\left\{ \operatorname{IoU}\left( \hat{\mathbf{b}}_{i}
    , \mathbf{b}_{i}\right) \ge k \right\},
\end{equation*}
where $N$ indicates the number of ground-truth objects, and $m$ denotes the
standard image captioning metrics, including BLEU-4~\cite{papineni_bleu_2002},
METEOR~\cite{banerjee_meteor_2005},
Rouge~\cite{lin_rouge_2004},
and CIDEr~\cite{vedantam_cider_2015},
abbreviated as B-4, M, R, and C, respectively. $\mathds{1}$ represents the
indicator function that is set to 1 if the IoU value for the $i$-{th} box is
bigger than the threshold $k$, otherwise 0.
\vspace{-10pt}
\paragraph{Implementation Details.}
For model configurations, we follow the set-up in TOD3Cap~\cite{jin_tod3cap_2024} unless otherwise specified. We note that MTA shares the same architecture as TOD3Cap, with the exception of our proposed BLA and DCA mechanisms. We employ the commonly used BEVFormer-tiny~\cite{li_bevformer_2022} as the pretrained BEV perception module. The BEV captioning module incorporates either the pretrained Llama-3.2-1B~\cite{llama_team_ai__meta_llama_2024} or InternLM2~\cite{cai_et_al_internlm2_2024} as the LLM. Unless stated otherwise, all studies are conducted with Llama 3. In all experiments, models are trained for 10 epochs with a learning rate of $2\times10^{-4}$, keeping the Llama model frozen except for the adapter~\cite{zhang_llama-adapter_2024} parameters. Additionally, the pretrained BEVFormer baseline is further trained for 10 epochs to ensure a fair comparison.

\begin{table*}[!t]
    \centering
    \small
    \resizebox{0.75\linewidth}{!}{
    \begin{tabular}{l|c|cccc|cccc}
        \hline
        \multirow{2}{*}{Method} & \multirow{2}{*}{LM} & \multicolumn{4}{c|}{$m\text{@IoU=}0.25$} & \multicolumn{4}{c}{$m\text{@IoU=}0.5$} \\
        \cdashline{3-10}
    && C$\uparrow$ & B-4$\uparrow$ & M$\uparrow$ & R$\uparrow$ & C$\uparrow$ & B-4$\uparrow$ & M$\uparrow$ & R$\uparrow$ \\
        \hline
        Scan2Cap~\cite{chen_scan2cap_2021}* & Custom~\cite{chen_scan2cap_2021} & 60.6 & 41.5 & 28.4 & 58.6 & 62.5 & 39.2 & 26.4 & 56.5\\
        X-Trans2Cap~\cite{yuan_x-trans2cap_2022}* & Custom~\cite{yuan_x-trans2cap_2022} & 99.8 & 45.9 & 35.5 & 66.8 & 92.2 & 43.3 & 34.7 & 65.7\\
        Vote2Cap-DETR~\cite{chen_vote2cap-detr_2024}* & Custom~\cite{chen_vote2cap-detr_2024} & 110.1 & 48.0 & 44.4 & 67.8 & 98.4 & 46.1 & 41.3 & 65.1\\
        TOD3Cap~\cite{jin_tod3cap_2024} & Llama 3.2~\cite{llama_team_ai__meta_llama_2024}& 113.1 & 48.7 & 49.8 & 68.0 & 108.7 & 46.7 & 47.8 & 65.3 \\
        \textbf{MTA(Ours)} & Llama 3.2~\cite{llama_team_ai__meta_llama_2024} & \textbf{122.8} & \textbf{49.4} & \textbf{50.3} & \textbf{68.7} & \textbf{118.7} & \textbf{47.6} & \textbf{48.4} & \textbf{66.2}\\
        TOD3Cap~\cite{jin_tod3cap_2024} & InternLM2~\cite{cai_et_al_internlm2_2024} & 109.0 & 47.6 & 49.4 & 67.5 & 105.1 & 45.6 & 47.4 & 64.8\\
        \textbf{MTA(Ours)} & InternLM2~\cite{cai_et_al_internlm2_2024} & 117.0 & 48.6 & 49.9 & 68.2 & 113.0 & 46.6 & 47.8 & 65.4\\
        \hline
    \end{tabular}
    }
    \vspace{-1ex}
    \caption{\textbf{3D dense captioning performance comparison on the TOD3Cap dataset}. The best-performing method is highlighted in \textbf{bold}. $^*$Results from \cite{jin_tod3cap_2024}. $\uparrow$: Higher values are better. LM: Language Model. The proposed MTA significantly outperforms the baseline TOD3Cap method and other state-of-the-art approaches across all metrics. When using Llama 3.2 as the language model, MTA achieves an $8.6\%$ improvement on C@0.25 and a $9.2\%$ improvement on C@0.5 compared to TOD3Cap. MTA shows similar improvements in the InternLM2 setting. These improvements are consistent across other metrics at both IoU thresholds of 0.25 and 0.5. 
    }
    \label{tab:tod3}
    
\end{table*}
\begin{table*}[!t]
    \centering
    \small
    \resizebox{0.9\linewidth}{!}{
    \begin{tabular}{l|c|cc|ccccc|cc}
        \hline
        \multirow{2}{*}{Method} & \multirow{2}{*}{LM} & \multirow{2}{*}{NDS$\uparrow$} & \multirow{2}{*}{mAP$\uparrow$} & \multicolumn{5}{c|}{True Positive Metrics} & \multicolumn{2}{c}{Inference Complexity}\\
        \cdashline{5-11}
        &&&& mATE$\downarrow$ & mASE$\downarrow$ & mAOE$\downarrow$ & mAVE$\downarrow$ & mAAE$\downarrow$ & \# Params & FPS\\
        \hline
        BEVFormer~\cite{li_bevformer_2022} & - & 37.4 & 26.8 & 0.903 & 0.292 & 0.611 & 0.573 & 0.221 & 33.6 M & 4.1\\
        TOD3Cap~\cite{jin_tod3cap_2024} & Llama 3.2~\cite{llama_team_ai__meta_llama_2024} & 37.7 & 26.6 & \underline{0.895} & 0.290 & \textbf{0.584} & 0.570 & 0.219 & 33.6 M & 4.1\\
        \textbf{MTA(Ours)} & Llama 3.2~\cite{llama_team_ai__meta_llama_2024} & \textbf{38.9} & \textbf{27.9} & 
        \textbf{0.878} & \underline{0.285} & \underline{0.595} & \underline{0.541} & \underline{0.213}
        & 33.6 M & 4.1\\
        TOD3Cap~\cite{jin_tod3cap_2024} & InternLM2~\cite{cai_et_al_internlm2_2024} & 37.7 & 26.8 & 0.905 & 0.288 & 0.596 & 0.562 & 0.215 & 33.6 M & 4.1\\
        \textbf{MTA (Ours)} & InternLM2~\cite{cai_et_al_internlm2_2024} & \underline{38.4} & \underline{27.6} & 0.899 & \textbf{0.281} & 0.608 & \textbf{0.540} & \textbf{0.212} & 33.6 M & 4.1\\
        \hline
    \end{tabular}
    }
    \vspace{-1ex}
    \caption{\textbf{3D detection performance comparison on the nuScenes validation set}. 
    The best-performing method is highlighted in \textbf{bold}, 
    while the second-best method is indicated by an \underline{underline}. LM: Language Model. The proposed MTA significantly outperforms the TOD3Cap method. Specifically, with Llama 3.2 as the language model, MTA surpasses TOD3Cap with a $3.2\%$ improvement in NDS and a $4.9\%$ improvement in mAP. Similar gains are observed when using InternLM2 as the language model. Note that MTA has same architecture as the baseline BEVFormer during inference, and thus it does not introduce additional complexity at runtime.}  
    \label{tab:det}
    \vspace{-5pt}
\end{table*}
\begin{table}[!t]
    \centering
    \resizebox{\linewidth}{!}{
    \begin{tabular}{l|c|ccc|cc|cc}
        \hline
        \multirow{2}{*}{Method} & Overall & \multicolumn{3}{c|}{Object Frequency} & \multicolumn{2}{c|}{Weather} & \multicolumn{2}{c}{Time of Day} \\
        \cdashline{3-9}
        & mAP & Rare & Common & Frequent &  Sunny & Rainy & Day & Night\\
        \hline
        BEVFormer &  26.8 & 17.6 & 33.3 & 47.4 & 26.7 & 28.8 & 27.0 & 12.6\\
        TOD3Cap &   26.6 &  17.1 & 33.3 & 47.7 & 26.4  & 28.8 & 26.8 & 12.4\\
        \textbf{MTA (Ours)} & \textbf{27.9} &  \textbf{18.9} & \textbf{34.2} & \textbf{47.9} & \textbf{27.7} & \textbf{29.7} & \textbf{28.1} & \textbf{13.1}\\
        \hline
    \end{tabular}
    }
    \vspace{-1ex}
    \caption{\textbf{mAP comparison on the nuScenes validation set across object frequency, weather conditions, and time of day.} The proposed MTA achieves the best performance across all scenarios, demonstrating robust generalization. Notable improvements are observed for rare objects ($<2\%$ frequency) and nighttime conditions, where MTA outperforms TOD3Cap by $10.5\%$ and $5.6\%$, respectively. (Object frequency: rare $<2\%$, common $2\%$--$20\%$, frequent $>20\%$)
    }
    \label{tab:generalization}
    \vspace{-5pt}
\end{table}

\subsection{Main Results}
\label{sec:results}
\paragraph{3D Dense Captioning Results.}
We compare the performance of the TOD3Cap network with that of the same models trained with the proposed MTA, as well as other state-of-the-art methods: Scan2Cap~\cite{chen_scan2cap_2021} uses a message passing graph module to facilitate learning object relation features. X-Trans2Cap~\cite{yuan_x-trans2cap_2022} applies a teacher-student approach to transfer detailed appearance information from 2D images to 3D scenes. Vote2Cap-DETR~\cite{chen_vote2cap-detr_2024} employs a one-stage design with dual prediction heads that decode scene features into bounding boxes and captions.

As shown in \cref{tab:tod3}, the proposed MTA demonstrates consistent and significant improvement over TOD3Cap across all captioning metrics in both the Llama 3.2 and InternLM2 settings. In the Llama 3.2 setting, MTA boosts the performance of the TOD3Cap baseline by $9.7$ points ($8.6\%$) on C@0.25 and $10.0$ points ($9.2\%$) on C@0.5. These substantial gains in CIDEr scores suggest that the captions generated with MTA are more closely aligned with the ground-truth descriptions, in terms of both n-gram overlap and capturing the importance
and relevance of the words within the context of the scene. Furthermore, MTA achieves improvements of 0.7 points ($1.4\%$) on BLEU-4@0.25 and 0.9 points ($1.9\%$) on BLEU-4@0.5, indicating enhanced fluency and grammatical correctness of the generated captions. Similar improvements are also observed for METEOR and ROUGE metrics at both IoU thresholds of 0.25 and 0.5. In the InternLM2 setting, MTA also demonstrates consistent improvements, aligning with those observed in the Llama 3.2 setting. These results highlight the effectiveness of MTA in enhancing the quality, relevance, and coherence of the generated captions, regardless of the language model used.
\vspace{-10pt}
\paragraph{3D Detection Results.}
As shown in \cref{tab:det}, MTA consistently outperforms the baseline models across almost all detection metrics. In this comparison, BEVFormer can be considered as the vision-only baseline, TOD3Cap as the vanilla multi-task learning baseline, and our proposed MTA as the aligned multi-task learning approach. In the Llama 3 setting, compared to the BEVFormer baseline, which was trained with the sole objective of maximizing detection performance, MTA achieves a $4.1\%$ improvement in mAP. Similarly, MTA outperforms the TOD3Cap method, yielding a $4.9\%$ increase in mAP. It is worth noting that the original TOD3Cap paper does not report detection results~\cite{jin_tod3cap_2024}. Hence, we retrained their model to obtain comparative numbers, which show a slight improvement in NDS and a slight drop in mAP compared to BEVFormer. We attribute this to the lack of alignment mechanisms focusing on both tasks in the original TOD3Cap method. Similar performance improvements are also observed in the InternLM2 setting. 
\begin{figure*}
    \centering
    \includegraphics[width=.85\linewidth]{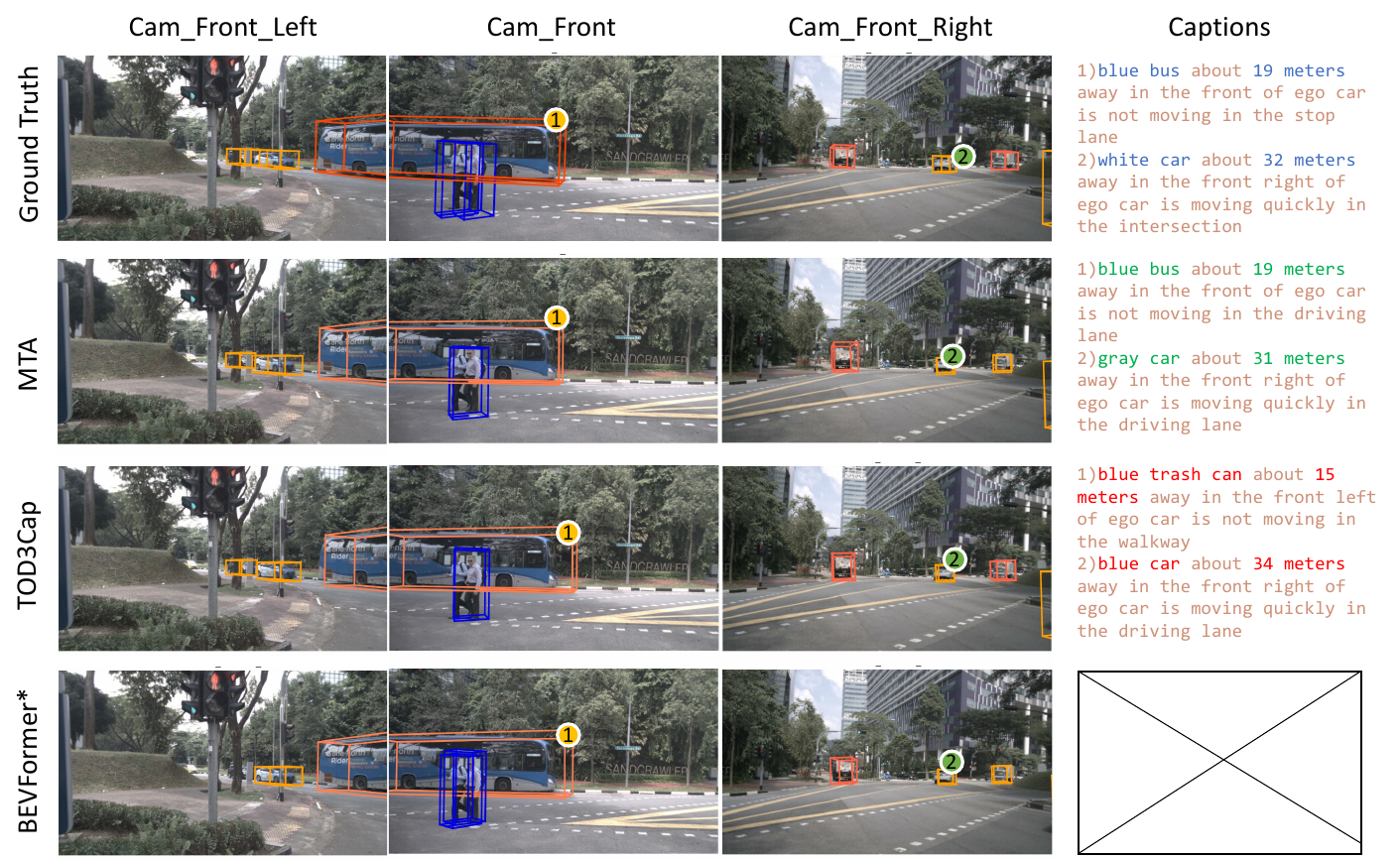}
    \caption{\textbf{Qualitative results comparing the proposed MTA with baseline methods on
    nuScenes and TOD3Cap datasets.} Visualization results show that MTA shows improved
    alignment with ground-truth detections over the counterpart methods.
    Captioning results show that the proposed MTA generates captions that are
    more accurate in terms of the description and localization of objects over the
    TOD3Cap. Unlike MTA, TOD3Cap labels object 1 (a bus) as a trash can in the
    caption, illustrating a heightened risk of hallucination. $^{*}$We note that
    BEVFormer is only suited for perception tasks, thus caption is not provided.
    }
    \label{fig:mta_visual_results}
    \vspace{-10pt}
\end{figure*}

\cref{tab:generalization} shows results on long-tail scenarios and challenging weather or day conditions. MTA demonstrates robust performance, achieving significantly improved performance over TOD3Cap with a $10.7\%$ increase in mAP for rare classes and a $5.7\%$ increase in nighttime conditions. These results highlight the effectiveness of the alignment mechanisms in MTA in enhancing the model's ability to handle long-tail scenarios.
\vspace{-10pt}
\paragraph{Computational Complexity.} One of the key advantages of MTA is that the BLA and DCA components are only used during training time. As a result, MTA does not require any architectural changes or introduce any additional complexity during inference, which is critical for real-world deployment. As shown in the last two columns of \cref{tab:det}, the number of parameters and frames per second (FPS) values at inference time are the same for all three methods (evaluated on a single V100 GPU), as the detection architecture and parameters are identical. This highlights the practical applicability of MTA, as it achieves improved performance without compromising on inference speed or model size.
\vspace{-10pt}
\paragraph{Qualitative Results.}
\cref{fig:mta_visual_results} presents a representative example to
visualize the qualitative results on the nuScenes and TOD3Cap validation sets. The detection results demonstrate that MTA achieves superior detection
quality and closer alignment with ground-truth detections compared to baseline methods.
Similarly, the captioning results highlight MTA's effectiveness in producing captions
that better match the ground-truth captions in terms of object description and localization
compared to the TOD3Cap baseline. For instance, 
the caption generated by the TOD3Cap method incorrectly identifies a bus as a
trash can, while MTA accurately describes the object. These qualitative results
demonstrate another crucial benefit of MTA's alignment mechanism: reducing the likelihood
of generating hallucinated captions, which is essential for safety-critical
applications like autonomous driving. The qualitative findings further confirm the
quantitative results, demonstrating the effectiveness of MTA in achieving
superior performance in both BEV perception and captioning tasks. Additional qualitative results are provided in the supplementary materials.
\subsection{Ablation Study}
\label{sec:ablation}
We perform extensive ablation studies to validate the key design choices in the proposed MTA framework. We note that additional ablation studies and comprehensive experimental tables are provided in the supplementary materials.

\noindent \textbf{BLA \& DCA Module Analysis.}
To gain a better understanding of the individual contributions of the proposed
BLA and DCA modules, we conduct an ablation study, as shown in
\cref{tab:ab_module}. We analyze the performance of each module separately and in
combination (MTA), comparing them to the baseline TOD3Cap model.

Notably, both the BLA and DCA mechanisms independently contribute to significant performance improvements over TOD3Cap. The BLA module has a particularly pronounced effect on detection performance, as evidenced by substantial increases in NDS and mAP scores. This finding suggests that aligning BEV features with linguistic representations enhances the model’s ability to accurately localize and classify objects within the 3D scene. In contrast, the DCA module has a stronger influence on captioning performance. This implies that enforcing consistency between detection and captioning outputs within the shared prompt space facilitates the generation of more accurate captions that are better grounded in the visual content. As a result, it reduces the occurrence of hallucinations or inconsistencies between the described objects and their spatial-temporal relationships within the scene. When both BLA and DCA modules are combined in our MTA framework, we achieve the best overall performance across all metrics, demonstrating the effectiveness of jointly optimizing 3D perception and captioning tasks through multimodal alignment. 

\begin{table}[!ht]
    \centering
    \resizebox{0.9\linewidth}{!}{
    \begin{tabular}{l|cc|cc}
        \hline
        \multirow{2}{*}{Method} & \multicolumn{2}{c|}{Captioning} & \multicolumn{2}{c}{Perception} \\
    \cdashline{2-5}       &  C@0.5$\uparrow$   & B-4@0.5$\uparrow$ & NDS$\uparrow$ & mAP$\uparrow$    \\
        \hline
        TOD3Cap         & 108.7             & \underline{46.7}  & 37.7          & 26.6             \\ 
        + BLA (Ours)                & 111.9             & 46.4              & \textbf{38.9} & \underline{27.7} \\
        + DCA (Ours)                & \underline{113.6} & 46.4              & 38.7          & 27.5             \\
        + MTA (Ours)       & \textbf{118.7}    & \textbf{47.6}     & \textbf{38.9} & \textbf{27.9}    \\
        \hline
    \end{tabular}
    }
    \caption{\textbf{Ablation study on the contributions of the BLA and DCA
    mechanisms.} Both the BLA and DCA mechanisms independently enhance overall performance
    over the TOD3Cap baseline. Combining both modules in the MTA framework
    yields the highest performance across all metrics. }
    \label{tab:ab_module}
    \vspace{-20pt}
\end{table}

\paragraph{Effect of $\ell$.}
We also investigate the impact of attaching the BLA objective to different
layers of the Q-Former, where the total number of layers is $L=8$. The results
in \cref{tab:ab_ell} demonstrate that aligning at the middle layer yields the
best performance. We hypothesize that aligning at an early stage forces the detection
embedding to directly mimic the text embedding without sufficient interaction with
the BEV features, potentially hindering the detection performance. Conversely,
aligning at a later stage of the Q-Former leaves little room for the remaining
layers to map the query to the MLLM space, considering that the text embedding from
the text encoder $\mathcal{T}$ differs from the MLLM, which can impede the
captioning performance. Consequently, in other experiments, we apply BLA to align
the first half of the Q-Former layers, striking a balance between allowing the
detection embedding to interact with the BEV features and providing sufficient
capacity for mapping to the MLLM space.

\begin{table}[ht]
    \centering
    \resizebox{.9\linewidth}{!}{
    \begin{tabular}{l|cc|cc}
        \hline
        \multirow{2}{*}{BLA Layer $\ell$} & \multicolumn{2}{c|}{Captioning} & \multicolumn{2}{c}{Perception} \\
    \cdashline{2-5}& C@0.5$\uparrow$   & B-4@0.5$\uparrow$ & NDS$\uparrow$    & mAP$\uparrow$    \\
        \hline
        1 (first)              & \underline{111.2} & \textbf{46.4}     & \underline{38.4} & \underline{27.3} \\ 
        4 (middle)          & \textbf{111.9}    & \textbf{46.4}     & \textbf{38.9}    & \textbf{27.7}    \\ 
        8 (last)                       & 107.7             & 45.8              & 37.9             & 27.1             \\
        \hline
    \end{tabular}
    }
    \caption{\textbf{Ablating the BLA attachment layer $\ell$.} Aligning at the middle
    layer achieves the best performance.}
    \label{tab:ab_ell}
    \vspace{-20pt}
\end{table}

\paragraph{Alignment Objective.}
We compare three training objectives for BLA and DCA alignment modules: Mean Squared Error (MSE),
Negative Cosine Similarity (Cos. Sim.), and the CLIP Contrastive Loss (CLIP)~\cite{radford_learning_2021}.

As shown in \cref{tab:ab_bla_loss}, we empirically find that for the BLA objective,
the MSE and CLIP objectives offer advantages over the Cos. Sim., with MSE showing
marginal advantages over the CLIP loss across all captioning and detection metrics.
Therefore, we opt for MSE as the objective for BLA. In contrast, for the DCA objective,
as presented in \cref{tab:ab_dca_loss}, the CLIP objective consistently outperforms
MSE and Cos.~Sim. across all metrics.

The difference in optimal objectives for BLA and DCA arises from their distinct alignment
goals and the nature of the modalities being aligned. BLA focuses on aligning
BEV features with linguistic representations, where the direct correspondence enforced
by MSE proves effective. In contrast, DCA aims to align detection and captioning
outputs within a shared prompt space, where the key is to establish
correspondences between information in the two modalities. For example, consider
the caption, ``traffic cone about 7 meters away in the back left of the ego car."
Here, the detection output should align the bounding box location with the
spatial description ``7 meters away in the back left," while the classification
label should correspond with the object category ``traffic cone'' mentioned in the
caption. The shared prompt space, leveraging the contrastive learning objective,
proves effective.

\begin{table}[ht]
    \centering
    \resizebox{.9\linewidth}{!}{
    \begin{tabular}{l|cc|cc}
        \hline
        \multirow{2}{*}{BLA Objective} & \multicolumn{2}{c|}{Captioning} & \multicolumn{2}{c}{Perception} \\
    \cdashline{2-5}& C@0.5$\uparrow$   & B-4@0.5$\uparrow$ & NDS$\uparrow$    & mAP$\uparrow$    \\
        \hline
        CLIP                & \underline{110.6} & \textbf{46.4}     & \underline{38.8} & \underline{27.6} \\ 
        MSE              & \textbf{111.9}    & \textbf{46.4}     & \textbf{38.9}    & \textbf{27.7}    \\ 
        Cos. Sim.       & 110.3             & 46.3              & 38.5             & 27.3             \\ 
        \hline
    \end{tabular}
    }
    \caption{\textbf{Ablating the BLA objective.} Mean squared error (MSE) achieves
    the best overall performance.}
    \vspace{-15pt}
    \label{tab:ab_bla_loss}
\end{table}

\begin{table}[ht]
    \centering
    \resizebox{.9\linewidth}{!}{
    \begin{tabular}{l|cc|cc}
        \hline
          \multirow{2}{*}{DCA Objective} & \multicolumn{2}{c|}{Captioning} & \multicolumn{2}{c}{Perception} \\
    \cdashline{2-5}& C@0.5$\uparrow$   & B-4@0.5$\uparrow$ & NDS$\uparrow$    & mAP$\uparrow$ \\
        \hline
        CLIP                & \textbf{113.6}    & \textbf{46.4}     & \textbf{38.7}    & \textbf{27.5} \\ 
        MSE                         & 112.4             & 46.3              & 38.0             & 27.2          \\ 
        Cos. Sim.       & \underline{113.1} & \textbf{46.4}     & \underline{38.3} & \textbf{27.5} \\ 
        \hline
    \end{tabular}
    }
    \caption{\textbf{Ablating the DCA objective.} The CLIP loss achieves the best
    results across all metrics.}
    \vspace{-10pt}
    \label{tab:ab_dca_loss}

\end{table}

\section{Conclusion}
\label{sec:conclusion} In this paper, we introduced MTA, a novel multimodal task alignment framework that bridges the gap between BEV perception and captioning tasks, significantly enhancing performance on both tasks. MTA consists of two novel mechanisms: BEV-Language Alignment (BLA) and Detection-Captioning Alignment (DCA). BLA harnesses a multimodal contextual learning mechanism to align BEV-based
visual representations and scene understanding with ground-truth linguistic representations. DCA leverages a cross-modal prompting mechanism to align detection and captioning outputs. Through extensive quantitative and qualitative experiments, we demonstrated the effectiveness of MTA in achieving improved performance on both BEV perception and captioning tasks. Importantly, MTA's alignment mechanisms are active only during training, ensuring no additional computational cost during inference, which is a critical factor in autonomous driving applications. 

{
    \small
    \bibliographystyle{ieeenat_fullname}
    \bibliography{bib/ma,bib/ma2}
}

\end{document}